\documentclass[journal]{IEEEtran}
\usepackage{amsmath,amsfonts}
\usepackage{amssymb}
\usepackage{algorithmic}
\usepackage{algorithm}
\usepackage{array}
\usepackage{booktabs}
\usepackage[table]{xcolor}
\usepackage{multirow}
\usepackage{makecell}
\usepackage[caption=false,font=footnotesize]{subfig}
\usepackage{textcomp}
\usepackage{stfloats}
\usepackage{url}
\usepackage{verbatim}
\usepackage{graphicx}
\graphicspath{{figures/}}
\usepackage{cite}
\usepackage[hidelinks]{hyperref}
\newcommand{\na}{\textemdash}
\newcommand{\cmark}{\checkmark}
\newcommand{\algstage}[1]{%
  \vspace{0.2em}%
  \STATE \textbf{#1}%
}

\begin{document}
\bstctlcite{IEEEcontrol}

\title{TriVAL: A Tri-Validation Framework for Faithful Automatic Optimization Modeling}

\author{Ziyang Fang, JinXi Wang, Jinghui Zhong,~\IEEEmembership{Senior Member,~IEEE}, and Yew-Soon Ong,~\IEEEmembership{Fellow,~IEEE}%
\thanks{(Corresponding author: Jinghui Zhong.)}%
\thanks{Ziyang Fang, JinXi Wang, and Jinghui Zhong are with the School of Computer Science and Engineering, South China University of Technology, Guangzhou 510000, China (e-mail: 202520143514@mail.scut.edu.cn, 202511095364@mail.scut.edu.cn, jinghuizhong@scut.edu.cn).}%
\thanks{Yew-Soon Ong is with the Centre for Frontier AI Research, Agency for Science, Technology and Research, Singapore 138632, and also with the College of Computing and Data Science, Nanyang Technological University, Singapore 639798 (e-mail: asysong@ntu.edu.sg).}}



\maketitle

\begin{abstract}
Optimization modeling serves as the pivotal bridge between natural-language problem descriptions and optimization solvers, and remains a cornerstone for bringing operations research (OR) into real-world decision making.
Recent advances in large language models (LLMs) have driven significant progress in automatic optimization modeling. However, existing methods still lack explicit validation during the modeling process, allowing errors introduced in earlier stages to carry through the pipeline and ultimately reduce final modeling accuracy.
To address this challenge, we introduce TriVAL, a tri-validation framework that performs explicit validation at three stages of automatic optimization modeling: semantic specification, mathematical formulation, and code generation. At each stage, TriVAL follows a construct-validate-revise loop that assesses the current result against stage-specific criteria and revises it when needed. This design helps identify and correct errors before they accumulate across stages, helping preserve faithfulness throughout the modeling process.
To evaluate automatic optimization modeling on more challenging combinatorial problems, we further introduce NL4COP, a benchmark of 150 instances across 50 diverse problem types with more complex decision logic, more tightly coupled constraints, and more demanding modeling requirements than existing benchmarks. Experiments on NL4COP and established benchmarks show that TriVAL consistently outperforms state-of-the-art methods, with the largest gains on the most challenging problems.
\end{abstract}

\begin{IEEEkeywords}
Combinatorial optimization, large language models, mathematical programming, automatic optimization modeling, optimization model validation
\end{IEEEkeywords}

\section{Introduction}

\IEEEPARstart{O}{ptimization} modeling constitutes the essential conduit for translating real-world decision problems into formal optimization models; nevertheless, it has long been the primary bottleneck preventing the large-scale deployment of operations research (OR) by non-specialist users.
This bottleneck is particularly consequential in domains such as supply chain planning \cite{nagurney2021optimization}, production scheduling \cite{xavier2021learning}, and transportation \cite{schenekemberg2021two}, where the need for optimization modeling far exceeds the availability of expert modelers \cite{williams2013model}.
In practice, optimization problems are rarely presented as fully specified mathematical programs.
They are typically described in natural language, often with intricate decision logic, tightly coupled requirements, and assumptions that are left implicit.
Constructing a correct mathematical model from such descriptions requires far more than direct formalization---it demands abstraction, semantic understanding, and careful modeling judgment.
Advancing this capability matters both for research on automatic optimization modeling and for extending the practical reach of OR beyond expert modelers \cite{xiao2025optimizationsurvey,wang2025large}.

Automatic optimization modeling is difficult because correctness must be maintained across a sequence of dependent modeling stages: understanding the problem, formulating it, and generating code \cite{jiang2025llmopt,astorga2025autoformulation,mostajabdaveh2024optimization}. 
Each stage produces a result that constrains the next; a flaw at any earlier stage can propagate forward; later results may still be internally coherent while disconnected from the original problem. 
Thus, the true bottleneck in automatic optimization modeling is no longer just the generation of code, but the robust containment of errors across dependent modeling stages to preserve faithfulness to the original problem.

Recent large language models (LLMs) have significantly advanced automatic optimization modeling \cite{xiao2025optimizationsurvey,wang2025large,wasserkrug2025enhancing}.
As optimization modeling tasks become more complex, existing work has increasingly adopted multi-step strategies to improve modeling capability.
Current approaches can be broadly categorized into two main lines: learning-based methods and prompt-based methods.
Learning-based methods enhance optimization modeling capability through large-scale data synthesis and model training \cite{huang2025orlm,jiang2025llmopt,lu2025optmath,chen2025solverinformed}, whereas prompt-based methods improve inference-time modeling through multi-step decomposition and the incorporation of modeling knowledge \cite{wang2025tdag,xiao2024chainofexperts,astorga2025autoformulation,thind2025optimai,zhang2025or}.
Both directions have improved the quality of generated models.
However, while these approaches enhance multi-step generation, they typically treat intermediate results from earlier stages as inherently reliable, giving much less attention to explicitly validating results before they are used in later stages.
Without explicit validation, early hallucinations, misinterpretations, or faulty reasoning can easily embed themselves into stage results.
Because later stages directly build on these inputs, they merely translate already-flawed modeling decisions into code. The final program may thus execute perfectly yet solve the wrong problem.
To mitigate semantic drift, automatic optimization modeling must explicitly validate key stage results, catching defects before they become difficult to correct later.

Recognizing that maintaining faithfulness requires catching defects at each stage before they become difficult to correct later, we propose TriVAL, a dedicated tri-validation framework for automatic optimization modeling that organizes the modeling process around explicit validation of intermediate modeling results.
It introduces three validations targeting distinct stages of the modeling process: semantic, formulation, and code.
For each target, TriVAL follows a construct-validate-revise loop: it first constructs the current modeling result, then validates it against criteria tailored to the specific risk at that stage, and revises it when necessary before proceeding.
Through this design, TriVAL contains early errors before they propagate, safeguarding the correct mapping from problem to solution throughout the modeling process.

The detrimental impact of error propagation is largely masked in existing LP and MILP benchmarks, where problem logic and formulation requirements are relatively straightforward.
However, in complex combinatorial settings with greater modeling difficulty, errors in problem understanding and formulation become more pervasive, making explicit validation critically important.
To evaluate optimization modeling where faithfulness is difficult to preserve, we introduce NL4COP, a new benchmark comprising 150 instances across 50 diverse combinatorial problem types with more complex decision logic and greater modeling difficulty than existing benchmarks.
Experiments on NL4COP and established benchmarks show that TriVAL consistently outperforms state-of-the-art methods, with the largest gains on the most challenging tasks.

The main contributions of this work are threefold:
\begin{enumerate}
\item We propose TriVAL, a dedicated tri-validation framework for automatic optimization modeling that organizes the modeling process around explicit validation of three key modeling results: semantic specification, mathematical formulation, and generated code. Each result is evaluated and, when necessary, revised before the process proceeds, helping contain error propagation and preserve faithfulness throughout the modeling process.

\item We introduce NL4COP, a new benchmark for automatic optimization modeling on challenging combinatorial problems. It comprises 50 problem types and 150 instances with complex decision logic and tightly coupled constraints, providing a more challenging and discriminative evaluation than existing benchmarks.

\item Through extensive experiments on NL4COP and established benchmarks, we show that TriVAL consistently outperforms state-of-the-art optimization-modeling methods, with the largest gains on the most challenging problems. The results further highlight NL4COP as a challenging and discriminative benchmark for demonstrating the benefits of explicit validation.
\end{enumerate}

The remainder of this paper is organized as follows. Section~\ref{sec:related_work} reviews related work and clarifies our positioning. Section~\ref{sec:methodology} presents the proposed TriVAL framework. Section~\ref{sec:experiments} introduces the NL4COP benchmark and reports the experimental evaluation. Finally, Section~\ref{sec:conclusion} concludes the paper.

\section{Related Work}
\label{sec:related_work}

Automatic optimization modeling translates natural-language problem descriptions into formal formulations and executable code for mathematical modeling frameworks and solvers such as CVXPY \cite{diamond2016cvxpy}, Gurobi \cite{gurobi2025}, and OR-Tools \cite{ortools2024}.
Recent literature applying LLMs to this task generally pursues two main directions.
The first direction trains specialized models via data synthesis, supervised fine-tuning, or alignment training.
The second direction guides general-purpose models using prompt engineering, reasoning frameworks, or external knowledge injection.
After reviewing these foundational approaches, we discuss the literature most closely aligned with our contribution: the evaluation of generated outputs and explicit optimization validation.

\subsection{Learning-Based Methods for Optimization Modeling}

Learning-based methods primarily advance automatic optimization modeling via supervised fine-tuning or alignment training, leveraging optimization-oriented data and feedback.
One prominent line of work focuses on constructing structured datasets to teach models to produce accurate formulations.
For instance, ORLM \cite{huang2025orlm} and LLMOPT \cite{jiang2025llmopt} design multi-element schemas that link natural-language descriptions, mathematical formulations, and code, enabling instruction tuning for optimization modeling.
Alternatively, ReSocratic \cite{yang2025optibench} and OptMATH \cite{lu2025optmath} employ formulation-centered synthesis, generating data bidirectionally between mathematical expressions and problem descriptions to expand training diversity with controllable complexity.

A complementary direction incorporates verifiable solver signals directly into model alignment and inference.
SIRL \cite{chen2025solverinformed} treats optimization solvers as reward sources, using executable code and formal models to guide reinforcement-learning-based alignment.
Building upon this verifiable feedback loop, OptiMind \cite{zhang2025optimind} applies solver signals at test time, combining class-based error summaries with self-consistency to reduce common modeling mistakes.

While these learning-driven approaches successfully produce higher-quality formulations and executable code, they remain fundamentally generation-focused.
They tend to treat results from earlier stages as inherently reliable, lacking explicit mechanisms to examine key modeling results before they are used in later stages.

\subsection{Prompt-Based Methods for Optimization Modeling}

A second major direction guides inference-time optimization modeling via task decomposition, multi-agent coordination, and structured search.
Given that single-step generation often fails on complex problem structures \cite{tsouros2023holy}, one prominent line of work decomposes the modeling workflow into specialized roles.
For instance, frameworks like Chain-of-Experts \cite{xiao2024chainofexperts}, OptiMUS \cite{ahmaditeshnizi2024optimus}, OptimAI \cite{thind2025optimai}, and OR-LLM-Agent \cite{zhang2025or} coordinate distinct agents or personas to separately handle mathematical formulation, code generation, and execution repair.
By modularizing the workflow, these methods allow models to focus on one specific part of the modeling task at a time.

A complementary line of work enforces structural consistency through search trajectories and semantic grounding.
OptiTree \cite{liu2025optitree} organizes the generation process via hierarchical tree search, decomposing complex problems into simpler subproblems to synthesize a global modeling strategy.
Alongside tree search, retrieving relevant modeling knowledge via in-context learning has shown promise for Constraint Programming \cite{michailidis2024constraint}.
To further ground the generation process, SAC-Opt \cite{zhang2025sacopt} focuses on structural alignment, reconstructing semantic anchors from the generated code and correcting mismatched components against the original problem description.

While these structural and multi-agent approaches make reasoning more reliable, their feedback loops are still concentrated at the code stage, typically through execution and debugging after code generation.
Because they rarely impose explicit validation gates on early modeling results (such as the initial semantic extraction or the mathematical formulation), early misinterpretations can still freely propagate into the final code structure.

\subsection{Evaluation of Generated Outputs and Explicit Optimization Validation}

Recognizing the limitations of pure generation, recent machine learning literature increasingly leverages LLMs as evaluators to critique and refine outputs \cite{madaan2023selfrefine,lightman2023let,chen2024teaching,gou2024critic,wang2026chief}.
Within optimization modeling, recent work has explored validation-guided search and testing.
For instance, AutoFormulation \cite{astorga2025autoformulation} incorporates correctness scoring to explore and screen candidate formulations.
Other works focus on post-generation verification: OptiVer \cite{liu2026optiver} checks final models via dual-side verification of structure and solutions, while an agent-based approach formalizes model checking using generated tests and mutations \cite{zadorojniy2025agent}.

Despite these advances, existing optimization verifiers are typically applied only at specific points of the modeling process or after a complete model has already been produced.
Because they do not systematically gate the intermediate transitions connecting natural language, mathematical formulations, and code, they struggle to prevent early semantic misunderstandings from cascading into complex formulation errors.

\begin{table}[t]
\caption{Comparison of representative optimization modeling methods. TriVAL proposes a dedicated tri-validation framework covering all core modeling stages.}
\label{tab:method-comparison}
\centering
\small
\setlength{\tabcolsep}{4pt}
\renewcommand{\arraystretch}{1.15}
\begin{tabular*}{\columnwidth}{@{\extracolsep{\fill}}lcccc@{}}
\toprule
Method & SU & F & CG & Validation Scope \\
\midrule
ORLM \cite{huang2025orlm}             &        & \cmark & \cmark & -- \\
LLMOPT \cite{jiang2025llmopt}           &        & \cmark & \cmark & -- \\
SIRL \cite{chen2025solverinformed}             &        & \cmark & \cmark & -- \\
AutoFormulation \cite{astorga2025autoformulation}  & \cmark & \cmark & \cmark & Formulation stage only \\
OptiMUS \cite{ahmaditeshnizi2024optimus}          & \cmark & \cmark & \cmark & -- \\
OR-LLM-Agent \cite{zhang2025or}     &        & \cmark & \cmark & -- \\
OptiTree \cite{liu2025optitree}         & \cmark & \cmark & \cmark & -- \\
OptiVer \cite{liu2026optiver}          & \cmark & \cmark & \cmark & Final-model only \\
\textbf{TriVAL (ours)} & \textbf{\cmark} & \textbf{\cmark} & \textbf{\cmark} & \textbf{Tri-Validation} \\
\bottomrule
\end{tabular*}

\vspace{2pt}
\footnotesize
SU: Semantic Understanding; F: Formulation; CG: Code Generation.
\end{table}

\subsection{Our Positioning}

As shown in Table~\ref{tab:method-comparison}, prior work covers the three broad stages of optimization modeling to different extents, while explicit validation is absent in most methods or introduced only partially.
TriVAL differs fundamentally: it elevates validation from a supporting component to the central mechanism of the framework.
Rather than introducing checks only at selected stages of the modeling process, TriVAL introduces explicit validation at each of the three stages: semantic understanding, formulation, and code generation.
By governing whether intermediate modeling results should be accepted, revised, or prevented from affecting later stages, this design contains error propagation at each stage, keeping the final generated code faithful to the original problem.

\section{Methodology: TriVAL}
\label{sec:methodology}
\subsection{Overview: A Tri-Validation Framework for Automatic Optimization Modeling}

TriVAL organizes automatic optimization modeling around explicit validation at three stages: semantic specification, mathematical formulation, and code generation. These stages produce the semantic specification $S$, the mathematical formulation $M$, and the generated code $C$, respectively. An error at any stage can carry forward and cause the final result to deviate from the original problem.

To address this risk, TriVAL introduces three validation gates, each targeting one stage of the modeling process. Semantic validation evaluates whether $S$ faithfully captures the original problem. Formulation validation examines whether $M$ correctly formalizes the specification. Code validation assesses whether $C$ faithfully translates $M$ and produces correct solutions.

Across all three stages, TriVAL follows a unified construct-validate-revise loop. The framework first constructs the current stage result, then evaluates it against stage-specific criteria, and revises it when needed before the modeling process moves forward. In this way, validation is embedded throughout the modeling process and acts as the central mechanism for containing error propagation.

To improve candidate quality before validation, TriVAL adopts stage-specific construction mechanisms: semantic extraction and ambiguity resolution for $S$, multi-expert formulation exploration for $M$, and a ReAct-based code agent with code self-correction for $C$. These mechanisms improve the quality of the results entering validation, making the overall process more reliable.

\begin{figure*}
\centering
\includegraphics[width=\textwidth]{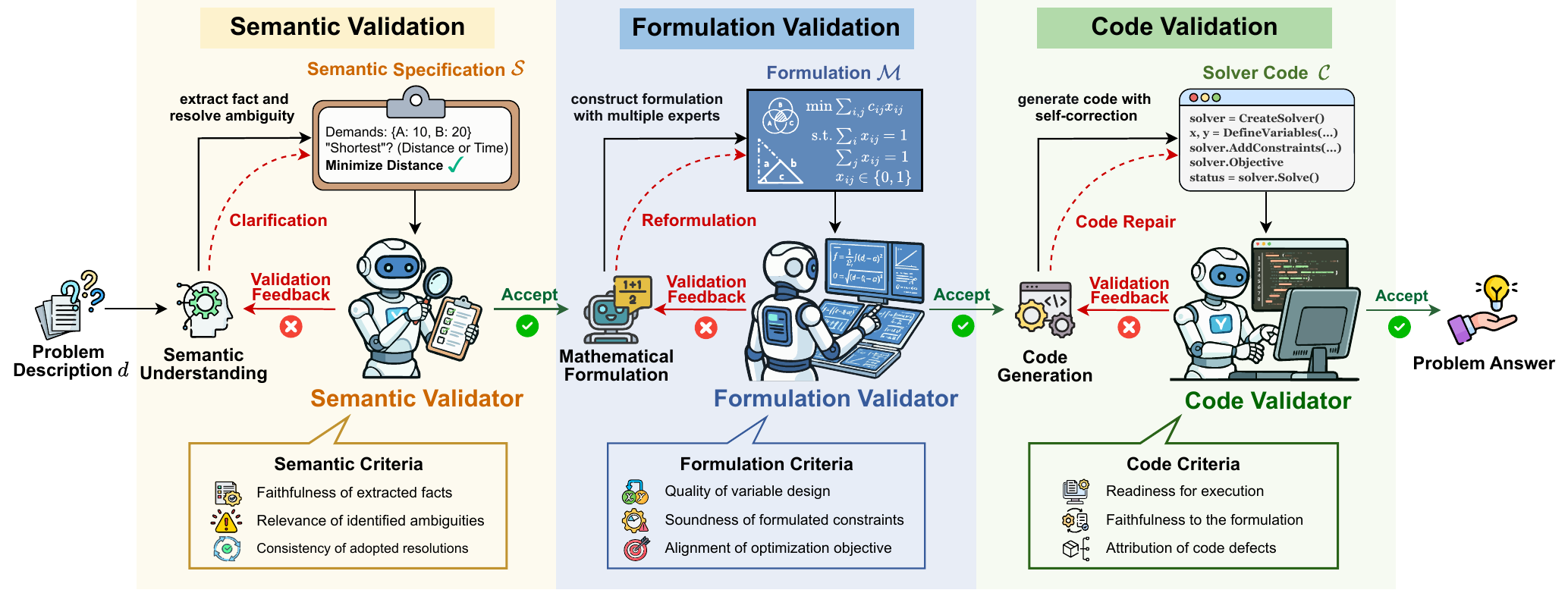}
\caption{Overview of TriVAL. The framework organizes automatic optimization modeling around three validation gates for the semantic specification $S$, mathematical formulation $M$, and generated code $C$. At each stage, construction, validation, and revision are performed before the process moves to the next stage.}
\label{fig:trival-framework}
\end{figure*}

Fig.~\ref{fig:trival-framework} illustrates the overall architecture of TriVAL, and Algorithm~\ref{alg:trival_progressive} summarizes the complete construct-validate-revise procedure.
The following subsections describe the three validation stages in detail.
Section~\ref{sec:experiments} further presents a case study demonstrating how TriVAL identifies and corrects modeling errors in practice.

\begin{algorithm}
\caption{TriVAL: The construct-validate-revise procedure}
\label{alg:trival_progressive}
\footnotesize
\begin{algorithmic}[1]
\STATE \textbf{Input:} problem description $d$
\STATE \textbf{Budgets:} semantic validation rounds $T_S$, formulation validation rounds $T_M$, code validation rounds $T_C$, code self-correction rounds $E_C$
\STATE \textbf{Output:} semantic specification $S$, mathematical formulation $M$, generated code $C$

\algstage{Semantic Validation}
\STATE $S \gets \textsc{ConstructSemanticSpecification}(d)$
\FOR{$t_S = 1$ to $T_S$}
    \STATE $(r_S, \delta_S) \gets \textsc{ValidateSemanticSpecification}(d, S)$
    \IF{$r_S = \texttt{accept}$}
        \STATE \textbf{break}
    \ELSIF{$r_S = \texttt{revise}$}
        \STATE $S \gets \textsc{ReviseSemanticSpecification}(S, \delta_S)$
    \ENDIF
\ENDFOR

\algstage{Formulation Validation}
\STATE $\mathcal{B} \gets \textsc{ConstructFormulationCandidates}(S)$
\STATE $M \gets \textsc{SelectFormulation}(\mathcal{B}, S)$
\FOR{$t_M = 1$ to $T_M$}
    \STATE $(r_M, \delta_M) \gets \textsc{ValidateFormulation}(d, S, M)$
    \IF{$r_M = \texttt{accept}$}
        \STATE \textbf{break}
    \ELSIF{$r_M = \texttt{partial\_revise}$}
        \STATE $M \gets \textsc{ReviseFormulation}(S, M, \delta_M)$
    \ELSIF{$r_M = \texttt{reformulate}$}
        \STATE $\mathcal{B} \gets \textsc{ConstructFormulationCandidates}(S, \delta_M)$
        \STATE $M \gets \textsc{SelectFormulation}(\mathcal{B}, S)$
    \ENDIF
\ENDFOR

\algstage{Code Validation}
\STATE $C \gets \textsc{GenerateSolverCode}(d, M)$
\FOR{$t_C = 1$ to $T_C$}
    \FOR{$e = 1$ to $E_C$}
        \STATE $(\eta_C, \phi_C) \gets \textsc{ExecuteCode}(C)$
        \IF{$\eta_C = \texttt{executable}$}
            \STATE \textbf{break}
        \ENDIF
        \STATE $C \gets \textsc{SelfCorrectCodeFromExecution}(C, \phi_C)$
    \ENDFOR
    \IF{$\eta_C \neq \texttt{executable}$}
        \STATE \textbf{break}
    \ENDIF
    \STATE $(r_C, \delta_C) \gets \textsc{ValidateSolverCode}(d, M, C, \phi_C)$
    \IF{$r_C = \texttt{accept}$}
        \STATE \textbf{return} $(S, M, C)$
    \ELSIF{$r_C = \texttt{code\_revise}$}
        \STATE $C \gets \textsc{ReviseSolverCode}(C, \delta_C)$
    \ELSIF{$r_C = \texttt{formulation\_revise}$}
        \STATE $M \gets \textsc{PartialReviseFormulation}(S, M, \delta_C)$
        \STATE $C \gets \textsc{GenerateSolverCode}(d, M)$
    \ENDIF
\ENDFOR

\STATE \textbf{return} $(S, M, C)$
\end{algorithmic}
\normalsize
\end{algorithm}

\subsection{Semantic Validation for Problem Understanding}

\begin{figure}[t]
\centering
\includegraphics[width=0.85\columnwidth]{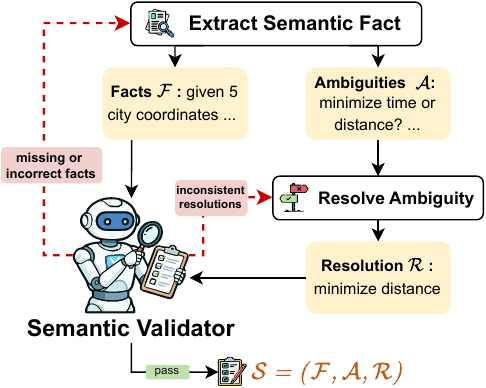}
\caption{Semantic validation targets the semantic specification $S=(\mathcal{F},\mathcal{A},\mathcal{R})$. $S$ is constructed through semantic fact extraction and ambiguity resolution, then validated for factual faithfulness, ambiguity relevance, and resolution consistency.}
\label{fig:semantic-validation}
\end{figure}

Semantic validation is the first validation gate in TriVAL, evaluating whether the problem has been understood faithfully at the semantic level before mathematical formulation begins.
This stage is critical because the semantic specification $S$ serves as the foundation on which all subsequent modeling depends: formulation inherits its starting point from $S$, and code generation in turn builds upon the formulation.
A specification that omits essential problem facts, introduces unsupported interpretations, or leaves formulation-relevant ambiguities unresolved can cause subsequent modeling to proceed from a mistaken understanding of the original problem, even if later stages remain internally consistent.

The specification $S$ under validation is a triple
\begin{equation}
S=(\mathcal{F},\mathcal{A},\mathcal{R}),
\end{equation}
where $\mathcal{F}$ captures semantic facts extracted from the problem description $d$, including given conditions, constraint requirements, and the optimization objective;
$\mathcal{A}$ identifies formulation-relevant ambiguities, i.e., parts of the description that admit multiple plausible interpretations and may affect modeling decisions (such as variable domains, parameter definitions, or the scope of constraints);
and $\mathcal{R}$ records the interpretations and conventions adopted for subsequent modeling.
Together, they capture the problem's given information, ambiguities, and adopted interpretations.
These ambiguities often affect the mathematical formulation directly, such as variable domains or the scope of constraints. Without resolving them, the formulation stage may adopt incorrect variable definitions or constraint scopes, introducing errors that are difficult to detect later.

TriVAL constructs $S$ through a two-step extraction-resolution process (Fig.~\ref{fig:semantic-validation}).
A semantic fact extractor first extracts $\mathcal{F}$ from the problem description and identifies candidate ambiguities $\mathcal{A}$.
An ambiguity resolver then determines the adopted interpretation for each ambiguity in $\mathcal{A}$ based on the problem context, the extracted facts, and standard optimization modeling conventions, producing the resolution set $\mathcal{R}$.
When no formulation-relevant ambiguity is identified, the extracted facts proceed directly to validation.
This specification provides the semantic basis for all subsequent modeling.

The semantic validator examines $S$ along three dimensions: factual faithfulness, ambiguity relevance, and resolution consistency.
For $\mathcal{F}$, it evaluates whether the extracted facts are complete, correct, and free of unsupported additions.
For $\mathcal{A}$, it gauges whether the listed ambiguities are genuinely relevant to formulation and sufficiently consequential to warrant resolution.
For $\mathcal{R}$, it determines whether the adopted resolutions are consistent with both the original problem description and the extracted facts.
These three dimensions jointly assess whether the current specification captures the semantic content of the original problem with sufficient faithfulness for formulation to proceed.

If the validator identifies defects, it returns \texttt{revise}, and TriVAL revises $S$ according to the validation feedback and re-evaluates it.
Missing or incorrect facts lead to revision of $\mathcal{F}$; missing, irrelevant, or misidentified ambiguities lead to revision of $\mathcal{A}$; and when the adopted interpretation is weakly supported or inconsistent with the problem description, TriVAL revises $\mathcal{R}$ accordingly.
This construct-validate-revise loop continues until the specification is accepted or the semantic validation round limit is reached, after which the current specification proceeds to the formulation stage.

\subsection{Formulation Validation for Mathematical Modeling}

\begin{figure}[t]
\centering
\includegraphics[width=0.85\columnwidth]{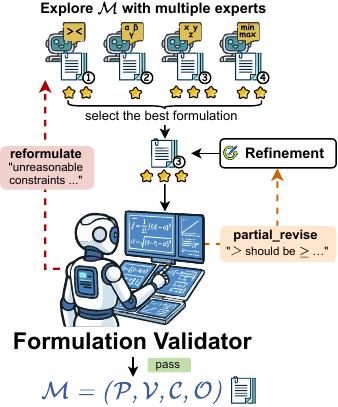}
\caption{Formulation validation targets the mathematical formulation $M=(P,V,\mathcal{C},O)$. Candidate formulations are constructed through multi-expert exploration, and the selected candidate is validated for variable-design quality, constraint soundness, and objective alignment.}
\label{fig:formulation-validation}
\end{figure}

With the semantic specification $S$ in place, the modeling process proceeds to mathematical formulation.
Formulation validation is the second validation gate, examining whether the problem has been correctly expressed as a mathematical formulation.
Because $M$ will be directly translated into code, any deviation from the specification propagates to subsequent stages.
Expression-level defects may prevent the model from being solved correctly or produce abnormal solver behavior. Modeling-level deviations are more consequential: they can cause the variables to misrepresent the intended decisions, the constraints to misstate the problem requirements, or the objective to no longer match the original goal, regardless of whether code generation itself is error-free.

The mathematical formulation is defined as
\begin{equation}
M=(P,V,\mathcal{C},O),
\end{equation}
where $P$ specifies the known parameters and constants in the model;
$V$ defines the decision variables that abstract the core choices in the optimization problem;
$\mathcal{C}$ encodes the constraints that feasible solutions must satisfy;
and $O$ expresses the optimization objective to be minimized or maximized.
Together, these four components constitute the complete mathematical representation of the problem and define the optimization model to be implemented.

To construct $M$, TriVAL employs multi-expert formulation exploration.
This design reflects that the same optimization problem can be approached from different modeling perspectives. Different perspectives lead to different choices of decision variables, which in turn produce different expressions of constraints and objectives, yielding formulations that are all mathematically valid yet vary substantially in compactness, solving difficulty, and ease of code generation.
For example, a routing problem may be formulated through an ordering-based representation or through an edge-based graph representation.
Both may be correct, yet they can differ substantially in variable design, constraint organization, and ease of code generation.
TriVAL uses multiple experts to explore these alternatives explicitly, so that validation can operate on a more diverse and higher-quality candidate set.

As illustrated in Fig.~\ref{fig:formulation-validation}, TriVAL explores candidate formulations through four experts.
Each expert takes the same semantic specification $S$ as input and independently produces a complete formulation candidate.
The four experts approach the problem from different modeling perspectives:

\begin{itemize}
\item Parameter-and-index expert: focuses on the organization of known quantities, emphasizing parameters, constants, and their index structure to provide a clear basis for later variable definition and constraint expression.
\item Decision-variable expert: focuses on decision representation, favoring variable designs that capture the core decisions compactly and sufficiently.
\item Constraint expert: focuses on requirement expression, emphasizing how constraints should be organized and scoped to express the problem requirements completely and accurately.
\item Objective expert: focuses on optimization-goal expression, favoring formulations whose objective is direct and well aligned with the variable design and constraint system.
\end{itemize}

This design allows TriVAL to explore different formulations of the same problem while keeping every candidate complete.
The exploration process yields a candidate set
\begin{equation}
\mathcal{B}={M_1,M_2,\dots,M_k}.
\end{equation}
From this candidate set, an LLM-based selector chooses the formulation that offers the best combination of sound variable design, concise representation, standard formulation conventions, and correctness with respect to $S$, and uses it as the formulation to be validated.

The formulation validator examines $M$ along three component-level dimensions:
\begin{equation}
\mathrm{Val}_M(d, S, M)=
\big(\mathrm{Qual}(V),\ \mathrm{Sound}(\mathcal{C}),\ \mathrm{Align}(O)\big),
\end{equation}
which correspond to the quality of variable design, the soundness of constraint formulation, and the alignment of the optimization objective.
The validator first checks variable design ($V$): decision variables should capture the core choices compactly and sufficiently, with domains, bounds, and types consistent with problem semantics.
It then examines constraint soundness ($\mathcal{C}$), focusing on completeness, correct scope, and accurate expression of problem requirements under the current variable design.
The final component-level check concerns objective alignment ($O$): the objective function should match the intended optimization goal in both direction and expression and remain coordinated with the variables and constraints.
Beyond these component-level dimensions, the validator also examines whether the formulation as a whole is coherent. Variables, constraints, and objectives are tightly coupled: variable design shapes how constraints and objectives are expressed, while constraint organization affects whether the overall formulation remains clear and well coordinated.

If the validator identifies defects, TriVAL determines the revision mode according to the nature of the defect.
Expression-level defects trigger \texttt{partial\_revise}.
These cases indicate that the overall formulation choice remains appropriate and that the main problem lies in local expressions, such as a missing constraint term, an improper variable bound, an incorrect scope condition, or a local deviation in the objective expression. Such defects can usually be corrected directly without changing the overall modeling design.
Modeling-level defects trigger \texttt{reformulate}.
These cases indicate that the defect lies in the overall formulation choice, such as an unsuitable decision abstraction, a constraint system that depends on an unsuitable variable design, or an objective that is misaligned with problem intent. In such cases, local patching is usually insufficient because variables, constraints, and objectives are tightly coupled, so changing one part can affect the rest of the formulation. TriVAL therefore returns to multi-expert formulation exploration and reconstructs candidate formulations under the same semantic specification, guided by the validation feedback. This step also allows the framework to revisit alternative modeling perspectives and potentially find a stronger formulation.
This construct-validate-revise loop continues until the formulation is accepted or the formulation validation round limit is reached, after which the formulation proceeds to code generation.

\subsection{Code Validation for Executable Translation}

\begin{figure}[t]
\centering
\includegraphics[width=0.85\columnwidth]{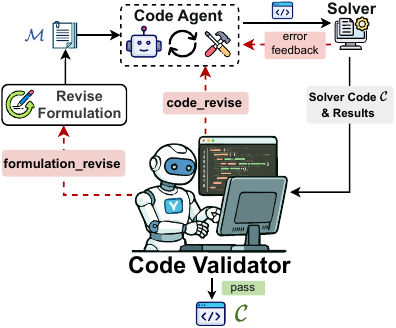}
\caption{Code validation targets the generated code $C$. The code agent generates code, and runtime errors are fixed through code self-correction. The resulting code is then validated for consistency with the formulation using the observed execution result. Detected defects are attributed to either the code or the formulation.}
\label{fig:code-validation}
\end{figure}

The formulation $M$ now enters the final stage: code generation.
Code validation is the third validation gate in TriVAL, assessing whether the formulation $M$ has been correctly translated into code $C$.
Two distinct error sources arise at this stage: solver-interface and code-expression errors, such as invalid variable types, incorrect constraint-construction calls, or syntax errors; and deviations from the formulation, where the code no longer accurately reflects the variables, constraints, and objective defined in $M$.
To address both sources, the code stage proceeds in two phases. Execution-feedback code self-correction first fixes runtime errors through execution feedback. Code validation then assesses whether the resulting code remains faithful to $M$.

To construct $C$, TriVAL introduces a ReAct-based code agent \cite{yao2023react} that constructs code iteratively by alternating reasoning and tool use.
The agent is equipped with three code-interaction tools for reading the current code state, writing initial code, and making partial edits to existing code.
These tools let the agent refine code statefully rather than regenerate the full program each time.
To improve adaptation to solver interfaces, TriVAL injects basic solver-interface syntax rules, common error patterns, and task guidance into the agent prompt as prior knowledge for code generation.

TriVAL pairs the code agent with a code self-correction loop: each time the code agent completes generation, an independent execution environment runs the code.
If execution fails, the environment returns execution feedback, including runtime errors, interface-call errors, and solver-status information.
This feedback is sent back to the code agent and used to guide the next round of partial revision.
This self-correction loop targets runtime and interface errors, iteratively fixing them through the returned feedback.
The loop continues until the code runs without errors or the repair budget is exhausted \cite{tang2024code}, returning an execution status $\eta_C$ together with execution feedback $\phi_C$.
Code self-correction resolves runtime errors but does not assess whether the resulting code remains faithful to $M$.
Code validation addresses this gap: even code that runs without errors may still return infeasibility or an abnormal solver status, and even code that produces a solution may still deviate from the formulation, yielding a solved problem that no longer matches $M$.

The code-stage validation is defined as
\begin{equation}
\mathrm{Val}_C(d,M,C,\phi_C)\to(r_C,\delta_C),
\end{equation}
where $\phi_C$ denotes the observed execution result and solver feedback, $r_C$ is the validation result, and $\delta_C$ carries the validation feedback, including error attribution and a description of the defect.

The validator inspects the executable code together with the observed execution result $\phi_C$, assessing faithfulness to $M$.
It checks that variable types, domains, and indices remain consistent with $M$, that constraints are added completely and correctly under the intended scope, and that the objective function remains aligned in direction and expression.
This faithfulness check also supports error attribution. If the code deviates from the formulation, the defect is attributed to the code, since code generation is responsible for translating mathematical expressions into executable solver code. If the code is consistent with $M$, the defect is attributed to the formulation. Certain formulation-level errors, such as incorrect constraint scope or missing coupling conditions, become visible only through actual execution when the solver result remains abnormal.
If the defect is attributed to the code (e.g., an interface-call error, a missing constraint term, an incorrect variable definition, or a deviation in objective expression), the validation result is \texttt{code\_revise}, and the code agent revises the current generated code accordingly.
If the defect is attributed to the formulation (e.g., a formulation defect exposed through infeasibility or an abnormal solver status), the validation result is \texttt{formulation\_revise}, the formulation is partially revised, and the code is regenerated from the revised formulation.
This construct-validate-revise loop continues until the code is accepted or the code validation round limit is reached, completing the TriVAL modeling process and producing the final results $(S, M, C)$.

\section{Experiments}
\label{sec:experiments}

This section evaluates TriVAL and NL4COP from four perspectives.
We first introduce NL4COP and the experimental setup.
We then assess the overall effectiveness of TriVAL and examine how its advantage changes with problem complexity across benchmarks and within NL4COP.
Next, we study the value of the complete validation framework through validator ablations, error analysis, and transfer experiments on an existing optimization-modeling framework.
Finally, we analyze how the three-stage modeling design and stage-specific mechanisms contribute to TriVAL's effectiveness, and evaluate the cost of validation.

\subsection{Experimental Setup}

\textbf{Benchmarks.} Existing benchmarks for automatic optimization modeling focus predominantly on LP and MILP problems with relatively simple structures \cite{michailidis2025cpbench}, while combinatorial optimization remains systematically underrepresented.
These problems involve complex constraint interactions and discrete decisions, making them particularly demanding to model.
To fill this gap, we introduce NL4COP, a benchmark that provides broad coverage of combinatorial problem types, graduated difficulty within each problem type, verified reference solutions, and strong discriminative power for distinguishing modeling methods.

NL4COP comprises 150 instances spanning 50 combinatorial problem types across seven major families: routing, packing and cutting, scheduling, location and allocation, graph and network optimization, knapsack and selection, and hybrid problems.
These 50 types are selected to systematically cover the principal modeling structures in combinatorial optimization, from path and flow decisions to resource coupling, assignment, sequencing, and set selection.

Each problem type contains three instances at distinct difficulty levels (simple, medium, and hard), enabling controlled comparison of modeling performance as difficulty increases within the same problem type.
The three levels differ in description length, constraint complexity, and data scale, placing progressively higher demands on long-context semantic understanding and mathematical formalization quality.

All instances are designed and constructed by PhD-level operations research experts, each grounded in a realistic OR scenario.
The problem data is verified for feasibility via solver execution, and every instance is fully specified in natural language with reference code and a reference optimal solution.
Two experts independently cross-check all cases for consistency among the problem description, reference code, and reference answer.

\begin{figure}[!t]
\centering
\includegraphics[width=\columnwidth]{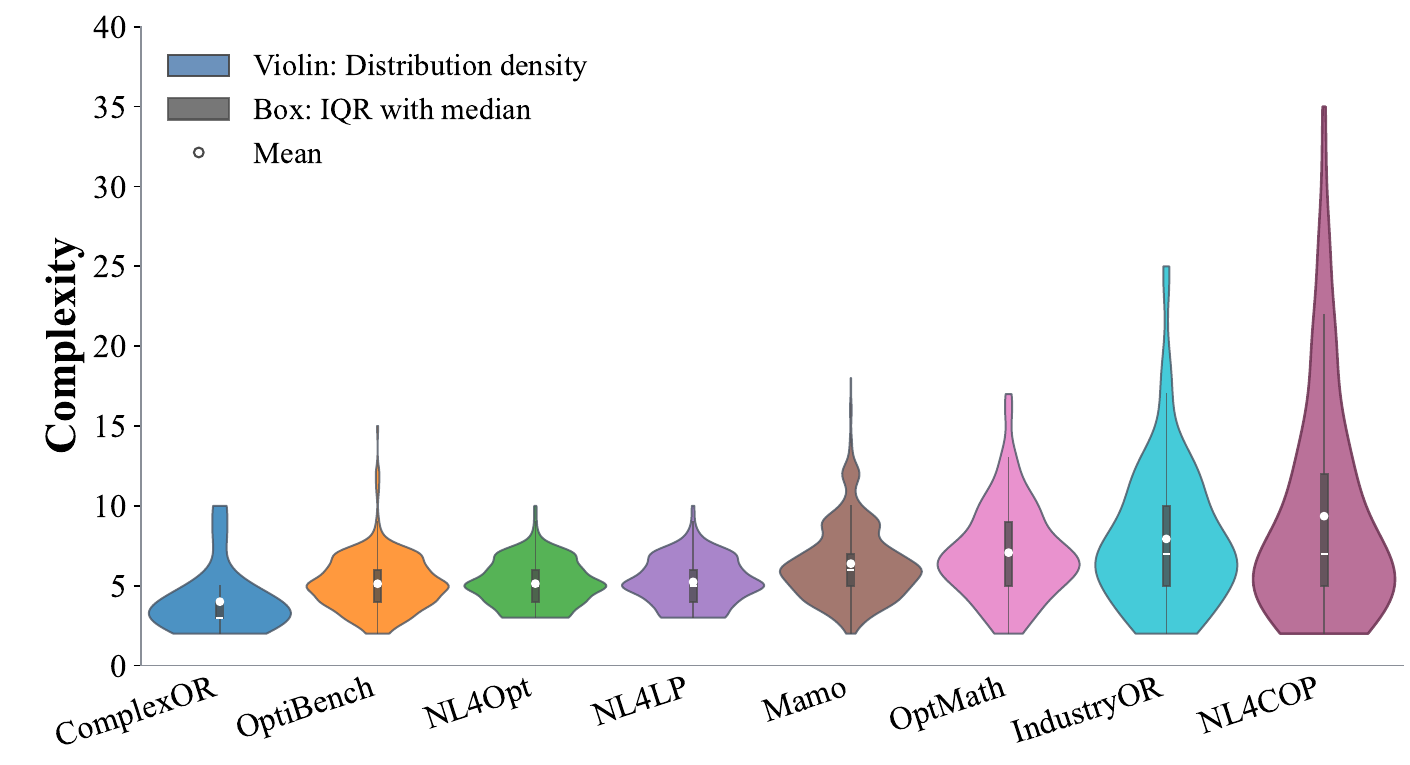}
\caption{Distribution of modeling complexity across benchmarks. NL4COP instances cluster at the high-complexity end, with substantially greater modeling difficulty than existing benchmarks.}
\label{fig:nl4cop-violin}
\end{figure}

Compared with existing benchmarks, NL4COP features more detailed problem descriptions and more intricate rule interactions, substantially increasing both description length and modeling complexity.
We compare NL4COP with existing benchmarks along these two dimensions, following the definition and computation of modeling complexity in prior work \cite{xiao2025optimizationsurvey}.
As shown in Table~\ref{tab:benchmark-complexity} and Fig.~\ref{fig:nl4cop-violin}, NL4COP substantially exceeds existing benchmarks in both dimensions.
Accordingly, constructing correct models for NL4COP requires more careful discrete abstraction, constraint scoping, and coordination across interacting constraints \cite{liu2026falcon}.

\begin{table}[!t]
\caption{Optimization modeling benchmarks compared by instance count, average description length, and modeling complexity. NL4COP ranks first in both description length and modeling complexity.}
\label{tab:benchmark-complexity}
\centering
\small
\renewcommand{\arraystretch}{1.18}
\setlength{\tabcolsep}{3pt}
\begin{tabular}{@{}>{\raggedright\arraybackslash}p{0.32\columnwidth}ccc@{}}
\toprule
Benchmark & Instances & Avg. Length (chars) & Complexity \\
\midrule
ComplexOR~\cite{xiao2024chainofexperts} & 18 & 1273 & 4.0 \\
OptiBench~\cite{yang2025optibench} & 403 & 621 & 5.1 \\
NL4Opt~\cite{ramamonjison2022nl4opt} & 205 & 530 & 5.1 \\
NL4LP~\cite{ahmaditeshnizi2024optimus} & 178 & 533 & 5.2 \\
Mamo~\cite{huang2025mamo} & 852 & 1225 & 6.4 \\
OptMath~\cite{lu2025optmath} & 129 & 3083 & 7.1 \\
IndustryOR~\cite{huang2025orlm} & 99 & 1046 & 7.9 \\
NL4COP (ours) & 150 & \textbf{3249} & \textbf{9.4} \\
\bottomrule
\end{tabular}
\end{table}

In addition to NL4COP, the main experiments include Mamo$_{\text{complex}}$ \cite{huang2025mamo}, OptMath \cite{lu2025optmath}, and IndustryOR \cite{huang2025orlm}.
Mamo$_{\text{complex}}$ is the complex subset of the Mamo benchmark, consisting of challenging LP and MILP instances.
OptMath spans a broader range of optimization problem families and includes longer descriptions with greater modeling difficulty.
IndustryOR focuses on optimization-modeling tasks drawn from real industrial settings.
These benchmarks, together with NL4COP, span established LP/MILP settings to more challenging combinatorial problems, enabling evaluation across varying modeling difficulty.

\textbf{Baselines.} We compare TriVAL against two classes of prior methods.
One class consists of learning-based methods, including ORLM (ORLM-LLaMA-3-8B) \cite{huang2025orlm}, LLMOPT (LLMOPT-Qwen2.5-14B) \cite{jiang2025llmopt}, and SIRL (SIRL-Gurobi32B) \cite{chen2025solverinformed}, which improve optimization modeling through training data, structured representations, and solver-verifiable learning signals.
The other consists of prompt- and agent-based methods, including AutoFormulation \cite{astorga2025autoformulation}, OR-LLM-Agent \cite{zhang2025or}, and OptiTree \cite{liu2025optitree}, which improve inference-time modeling through task decomposition, search, planning, external feedback, and multi-agent coordination.

\textbf{Evaluation Metric.} Following prior work \cite{huang2025orlm,jiang2025llmopt,liu2025optitree,chen2025solverinformed,astorga2025autoformulation}, we adopt solving accuracy as the primary metric.
No general-purpose automatic test of structural equivalence currently exists for optimization modeling, so solving accuracy remains the standard evaluation criterion.
For each instance, we execute the generated code, obtain the predicted objective value $y_{\text{pred}}$, and compare it with the reference objective value $y_{\text{label}}$.
Following the SIRL evaluation protocol \cite{chen2025solverinformed}, an instance is counted as correct when
\begin{equation}
\frac{|y_{\text{pred}}-y_{\text{label}}|}{|y_{\text{label}}|+1} < 10^{-6}.
\end{equation}
We also report process-level indicators and error-type analyses as supplementary evidence.
For the small number of instances where the decision-variable type is not explicitly specified, we evaluate both the integral and the continuous settings.
If either setting matches the reference objective value, we count the instance as correct.
This rule is applied uniformly across all methods.

\textbf{Protocol.} The main experiments use DeepSeek-V3.2 \cite{deepseekai2025deepseekv3} and GPT-5.1 \cite{openai2025gpt51announcement} as base models.
For prompt- and agent-based methods, TriVAL and all baselines are evaluated on the same base model to ensure a fair comparison.
For learning-based methods, we evaluate the authors' released models under their best reported settings.
For all prompt- and agent-based methods evaluated by us, we repeat each experiment five times.
Unless otherwise specified, we report the best solving accuracy over the five runs.
We use OR-Tools \cite{ortools2024} and CVXPY \cite{diamond2016cvxpy} as solvers to cover the optimization problem types in all benchmarks.
In TriVAL, each stage allows up to five validation iterations, after which the current result proceeds to the next stage.
The code self-correction stage allows up to 20 rounds of execution feedback, with a timeout of 100~s per run.
Unless a module is explicitly removed in an ablation, all other settings are kept fixed across variants.

\begin{table*}[!t]
\caption{Accuracy (\%) across benchmarks. TriVAL achieves the strongest overall performance, with the largest gains on the most challenging benchmarks.}
\label{tab:main-results}
\centering
\small
\setlength{\tabcolsep}{6pt}
\renewcommand{\arraystretch}{1.18}
\begin{tabular*}{0.94\textwidth}{@{\extracolsep{\fill}}llccccc@{}}
\toprule
\multirow{2}{*}{Model} & \multicolumn{1}{c}{\multirow{2}{*}{Method}} & \multicolumn{4}{c}{Benchmark Accuracy (\%)} & \multirow{2}{*}{Overall (\%)} \\
\cmidrule(lr){3-6}
& & Mamo$_{\text{complex}}$ & OptMath & IndustryOR & NL4COP & \\
\midrule
\rowcolor{gray!20}
\multicolumn{7}{c}{\textbf{Learning-based Methods}} \\
\midrule
\multirow{3}{*}{} & ORLM & 48.6 & 14.1 & 34.3 & 11.3 & 29.1 \\
& LLMOPT & 50.0 & 21.9 & 38.4 & 10.7 & 31.9 \\
& SIRL & 65.2 & 71.1 & 52.5 & 28.0 & 54.9 \\
\midrule
\rowcolor{gray!20}
\multicolumn{7}{c}{\textbf{Prompt-based Methods}} \\
\midrule
\multirow{4}{*}{GPT-5.1} & AutoFormulation & 74.3 & 46.9 & 52.5 & 25.3 & 52.1 \\
& OR-LLM-Agent & \textbf{91.0} & 85.2 & 78.8 & 68.0 & 81.8 \\
& OptiTree & 80.5 & 83.6 & 75.8 & 56.0 & 74.1 \\
& \textbf{TriVAL} & \underline{90.5} & \textbf{87.5} & \underline{86.9} & \underline{81.3} & \underline{86.9} \\
\midrule
\multirow{4}{*}{DeepSeek-V3.2} & AutoFormulation & 74.8 & 40.6 & 63.6 & 14.7 & 50.1 \\
& OR-LLM-Agent & 88.6 & \underline{85.9} & 77.8 & 52.0 & 76.8 \\
& OptiTree & 82.9 & 74.2 & 67.7 & 54.7 & 71.2 \\
& \textbf{TriVAL} & 90.0 & \textbf{87.5} & \textbf{89.9} & \textbf{87.3} & \textbf{88.8} \\
\bottomrule
\end{tabular*}

\vspace{2pt}
\footnotesize
Best results are in \textbf{bold} and second-best results are \underline{underlined}.
\end{table*}

\subsection{Main Results}

Table~\ref{tab:main-results} reports the solving accuracy of TriVAL and competing methods on NL4COP and three established optimization-modeling benchmarks.
TriVAL achieves the strongest overall performance across all benchmarks and both base models, with particularly large gains on IndustryOR and NL4COP.
The advantage is largest on IndustryOR and NL4COP because longer descriptions and greater modeling difficulty make modeling more error-prone, and methods without explicit validation allow these errors to propagate unchecked, whereas TriVAL's staged validation catches them early.

\begin{figure}[tbp]
\centering
\includegraphics[width=\columnwidth]{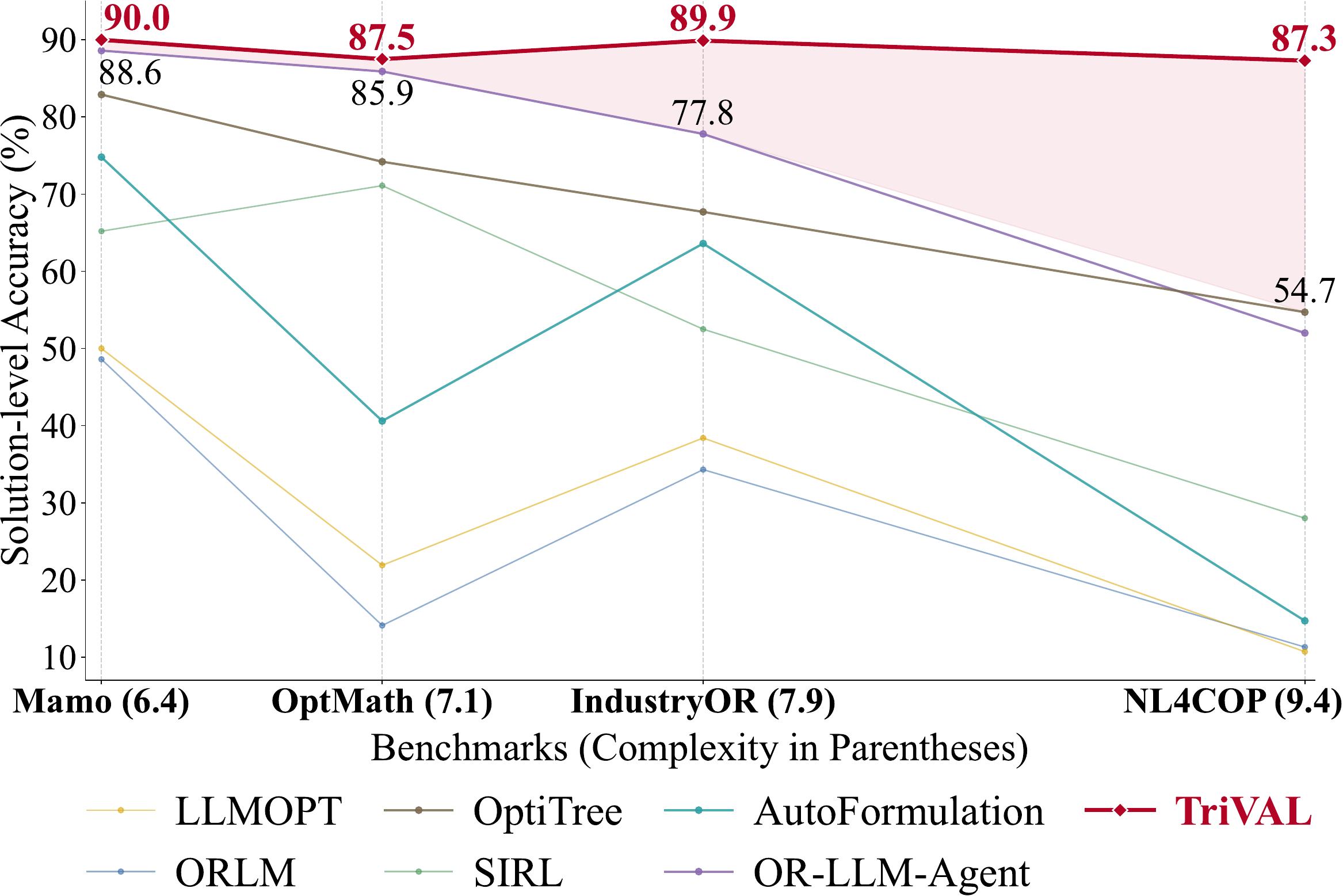}
\caption{Solving accuracy versus benchmark complexity under DeepSeek-V3.2. As benchmark difficulty increases, all methods decline but TriVAL degrades the slowest, widening the gap to competing methods.}
\label{fig:benchmark-performance-panels}
\end{figure}

Fig.~\ref{fig:benchmark-performance-panels} traces this trend across benchmarks.
As complexity increases, all methods decline, but TriVAL degrades much more slowly: under DeepSeek-V3.2, TriVAL drops only 2.7 points from Mamo$_{\text{complex}}$ to NL4COP, whereas OR-LLM-Agent drops 36.6 and OptiTree drops 28.2 (Table~\ref{tab:main-results}).
NL4COP thus provides strong discriminative power among methods: on simpler benchmarks, most approaches perform adequately and differences remain small, whereas NL4COP's greater complexity amplifies these differences, clearly separating methods with and without explicit validation.

Table~\ref{tab:nl4cop-difficulty} breaks down the NL4COP results by difficulty level.
Under both base models, the gap between TriVAL and the baselines widens from simple to hard splits, consistent with the cross-benchmark trend observed above.
This within-benchmark result further shows that TriVAL is more robust on harder instances, and that NL4COP's graduated difficulty reveals this difference.

\begin{table}[!t]
\caption{Solving accuracy across the simple, medium, and hard splits of NL4COP (\%). The gap between TriVAL and the baselines widens as difficulty increases.}
\label{tab:nl4cop-difficulty}
\centering
\small
\setlength{\tabcolsep}{2pt}
\renewcommand{\arraystretch}{1.18}
\begin{tabular}{@{}llcccc@{}}
\toprule
\multirow{2}{*}{Model} & \multirow{2}{*}{Method} & \multicolumn{3}{c}{Difficulty Splits} & \multirow{2}{*}{Overall} \\
\cmidrule(lr){3-5}
& & Simple & Medium & Hard & \\
\midrule
\multirow{3}{*}{GPT-5.1}
& OR-LLM-Agent & 72.0 & 72.0 & 60.0 & 68.0 \\
& OptiTree & 52.0 & 56.0 & 60.0 & 56.0 \\
& TriVAL & \underline{88.0} & \underline{90.0} & \underline{66.0} & \underline{81.3} \\
\midrule
\multirow{3}{*}{DeepSeek-V3.2}
& OR-LLM-Agent & 56.0 & 56.0 & 44.0 & 52.0 \\
& OptiTree & 60.0 & 58.0 & 46.0 & 54.7 \\
& TriVAL & \textbf{94.0} & \textbf{92.0} & \textbf{76.0} & \textbf{87.3} \\
\bottomrule
\end{tabular}
\end{table}

\subsection{Effectiveness of the Validation Framework}

\subsubsection{Validator Ablation}

To examine the role of each validation stage, we keep the construction process unchanged and remove the semantic, formulation, and code validators individually or jointly.

Table~\ref{tab:validation-ablation-industryor} reports the ablation results on IndustryOR.
Explicit validation improves accuracy by at least 11.1 percentage points (pp) and up to 13.1 pp, showing that it is essential to TriVAL's performance.
Among the three, the formulation validator contributes the most, improving accuracy by 4.0--8.1 pp, as it guards the critical interface where semantic understanding is translated into concrete mathematical expressions; errors at this stage directly corrupt code generation and are difficult to recover from in later stages.
The code validator improves accuracy by 2.0--6.1 pp by addressing more local implementation defects such as solver-interface mismatches, while the semantic validator contributes 1.0--4.0 pp by catching early misinterpretations that would otherwise cascade through both formulation and code.
Together, these results support the core design of TriVAL: the three validators target distinct failure modes across the modeling pipeline, and each addresses errors that the others do not catch.

We further examine the same ablation on NL4COP, where the greater modeling difficulty makes errors more likely to arise.
The effect of validation is substantially larger: the complete framework improves accuracy by at least 18.0 pp and up to 21.3 pp, with the formulation validator alone contributing 11.3--14.0 pp (Table~\ref{tab:validation-ablation-nl4cop}).
Every individual ablation produces a larger degradation than on IndustryOR, showing that validation becomes more valuable as modeling difficulty increases.
This growing gap also highlights NL4COP's discriminative power: the benchmark's complex decision logic and greater modeling difficulty amplify the difference between methods with and without validation, making the contribution of each validator clearly observable.

\begin{table}[!t]
\caption{Validation ablation on IndustryOR. All three validators contribute to the overall accuracy.}
\label{tab:validation-ablation-industryor}
\centering
\small
\setlength{\tabcolsep}{6pt}
\renewcommand{\arraystretch}{1.18}
\begin{tabular}{@{}lcc@{}}
\toprule
Variant & Accuracy (\%) & $\Delta$ (pp) \\
\midrule
\textbf{Full Method} & \textbf{89.9} & \na \\
w/o Semantic Validation & 85.9--88.9 & -1.0 -- -4.0 \\
w/o Formulation Validation & 81.8--85.9 & -4.0 -- -8.1 \\
w/o Code Validation & 83.8--87.9 & -2.0 -- -6.1 \\
w/o All Validation & 76.8--78.8 & -11.1 -- -13.1 \\
\bottomrule
\end{tabular}

\vspace{2pt}
\footnotesize
Accuracy (\%): min--max over five runs. $\Delta$: drop from Full Method in percentage points (pp).
\end{table}

\begin{table}[!t]
\caption{Validation ablation on NL4COP. The effect of each validator is larger than on IndustryOR, reflecting the greater modeling difficulty of this benchmark.}
\label{tab:validation-ablation-nl4cop}
\centering
\small
\setlength{\tabcolsep}{6pt}
\renewcommand{\arraystretch}{1.18}
\begin{tabular}{@{}lcc@{}}
\toprule
Variant & Accuracy (\%) & $\Delta$ (pp) \\
\midrule
\textbf{Full Method} & \textbf{87.3} & \na \\
w/o Semantic Validation & 82.0--86.0 & -1.3 -- -5.3 \\
w/o Formulation Validation & 73.3--76.0 & -11.3 -- -14.0 \\
w/o Code Validation & 78.7--84.0 & -3.3 -- -8.7 \\
w/o All Validation & 66.0--69.3 & -18.0 -- -21.3 \\
\bottomrule
\end{tabular}

\vspace{2pt}
\footnotesize
Accuracy (\%): min--max over five runs. $\Delta$: drop from Full Method in percentage points (pp).
\end{table}

\subsubsection{Reduction of Variable and Constraint Errors}

Having established that validation improves accuracy, we now ask what types of errors it actually fixes.
Table~\ref{tab:validation-error-distribution} classifies failed cases into three categories. Variable-design errors include wrong variable domains, missing decision variables, and incorrect index scopes. Constraint-expression errors include missing constraints, wrong scopes, and incorrect coupling relations. Code-generation errors involve solver-interface mismatches and other translation-level mistakes.
The key observation is that removing validation disproportionately increases errors in the first two categories.
On NL4COP, constraint errors more than double from 12 to 30 and variable errors double from 6 to 12, showing that TriVAL's validators mainly reduce errors in variable design and constraint formulation.
These are precisely the errors that undermine faithfulness to the original problem: wrong variable abstractions alter the decision space, and flawed constraint formulations distort the feasible region, regardless of whether the subsequent code executes correctly.
By catching these defects early, the three validators directly protect the key modeling stages, helping the solver optimize the intended problem.

\begin{table}[!t]
\caption{Error distribution on failed instances with and without validation. Removing validation mainly increases variable-design and constraint-expression errors rather than code-generation errors.}
\label{tab:validation-error-distribution}
\centering
\small
\setlength{\tabcolsep}{4pt}
\renewcommand{\arraystretch}{1.18}
\resizebox{\columnwidth}{!}{%
\begin{tabular}{@{}llcccc@{}}
\toprule
\multirow{2}{*}{Benchmark} & \multirow{2}{*}{Variant} & \multirow{2}{*}{\makecell[c]{Total\\failed}} & \multicolumn{3}{c}{Error Types} \\
\cmidrule(lr){4-6}
& & & Code & Variable & Constraint \\
\midrule
\multirow{2}{*}{IndustryOR}
& Full Method & \textbf{10} & 1 & \textbf{1} & \textbf{8} \\
& w/o All Validation & 21 & 1 & 3 & 17 \\
\midrule
\multirow{2}{*}{NL4COP}
& Full Method & \textbf{20} & 2 & \textbf{6} & \textbf{12} \\
& w/o All Validation & 46 & 4 & 12 & 30 \\
\bottomrule
\end{tabular}
}
\end{table}

\subsubsection{Applying Validation to an Existing Framework}

The preceding experiments show that explicit validation improves modeling accuracy within TriVAL. We now ask whether the same mechanism can benefit another optimization-modeling framework.
To test this, we add explicit validation into OR-LLM-Agent \cite{zhang2025or}, a representative multi-stage framework that performs optimization modeling through mathematical modeling, code generation, and debugging.
We keep its original modeling pipeline unchanged and insert formulation-side validation after the mathematical modeling stage and code-side validation after the debugging loop produces executable code.
Detected formulation-side defects trigger natural-language feedback to the mathematical-modeling stage; code-side defects trigger feedback to the code-generation stage, while runtime errors are still handled by the original debugging loop.
The added validators reuse the same prompts as in TriVAL, with each validator allowed up to five iterations.

After adding these validation steps, accuracy improves from 77.8\% to 83.2\% on IndustryOR and from 52.0\% to 70.2\% on NL4COP.
Formulation-side validation consistently detects more errors than code-side validation, flagging 37.4\% vs.\ 19.2\% of instances on IndustryOR and 42.7\% vs.\ 22.7\% on NL4COP, indicating that many modeling defects arise during the mathematical formulation stage.
On NL4COP, both validators trigger revision more frequently than on IndustryOR, and the accuracy gain is correspondingly larger, further reinforcing that validation grows more valuable as problem difficulty increases.
These results demonstrate that explicit validation can also improve a representative existing multi-stage framework.

\begin{table}[!t]
\caption{Effect of adding validation to OR-LLM-Agent. Validation improves accuracy on both benchmarks, with a substantially larger gain on NL4COP.}
\label{tab:orllmagent-validation}
\centering
\small
\setlength{\tabcolsep}{3.5pt}
\renewcommand{\arraystretch}{1.18}
\begin{tabular*}{\columnwidth}{@{\extracolsep{\fill}}lccc@{}}
\toprule
Benchmark & OR-LLM-Agent & + Validation & Gain \\
\midrule
IndustryOR & 77.8 & 83.2 (82.8--83.8) & +5.4 (5.0--6.0) \\
NL4COP & 52.0 & 70.2 (68.7--72.0) & +18.2 (16.7--20.0) \\
\bottomrule
\end{tabular*}

\vspace{2pt}
\footnotesize
OR-LLM-Agent: original mean accuracy. + Validation: mean (min--max) over five runs after inserting formulation-side and code-side validation.
\end{table}

Across these experiments, explicit validation consistently improves modeling accuracy, becomes more effective as problem difficulty increases, and also improves a representative existing framework. We next analyze the internal design choices that contribute to TriVAL's effectiveness.

\subsection{Analysis of TriVAL's Design}
\label{sec:further-analysis}

\subsubsection{Contribution of Modeling Design and Mechanisms}

The preceding section establishes that explicit validation is a key driver of TriVAL's performance. We now turn to the broader framework: how the three-stage modeling structure and the stage-specific mechanisms contribute to the overall effectiveness.
To isolate each component's role, we keep the validator configuration fixed and evaluate the following variants: without semantic understanding, without formulation, without both stages (code-only), without multi-expert formulation, and without code self-correction. All variants are evaluated on IndustryOR and NL4COP.

\begin{table*}[!t]
\caption{Ablation of modeling design choices on IndustryOR and NL4COP. Both the three-stage structure and the stage-specific mechanisms contribute to final accuracy, with the largest drops from removing semantic understanding or code self-correction.}
\label{tab:ablation-stage-mechanisms}
\centering
\small
\setlength{\tabcolsep}{4pt}
\renewcommand{\arraystretch}{1.18}
\resizebox{\textwidth}{!}{%
\begin{tabular}{@{}lccccc|cc|cc@{}}
\toprule
Variant & \makecell[c]{Semantic\\Understanding} & \makecell[c]{Formulation} & \makecell[c]{Code\\Generation} & \makecell[c]{Multi-Expert\\Formulation} & \makecell[c]{Code Self-\\Correction} & IndustryOR & $\Delta$ & NL4COP & $\Delta$ \\
\midrule
\multicolumn{10}{c}{\textbf{Three-Stage Structure}} \\
\midrule
\textbf{Full Method} & $\checkmark$ & $\checkmark$ & $\checkmark$ & $\checkmark$ & $\checkmark$ & \textbf{89.9} & \na & \textbf{87.3} & \na \\
w/o Semantic Understanding & $\times$ & $\checkmark$ & $\checkmark$ & $\checkmark$ & $\checkmark$ & 80.8 & -9.1 & 78.7 & -8.6 \\
w/o Formulation Stage & $\checkmark$ & $\times$ & $\checkmark$ & \na & $\checkmark$ & 83.8 & -6.1 & 78.0 & -9.3 \\
Code-Only & $\times$ & $\times$ & $\checkmark$ & \na & $\checkmark$ & 82.8 & -7.1 & 78.0 & -9.3 \\
\midrule
\multicolumn{10}{c}{\textbf{Stage-Specific Mechanisms}} \\
\midrule
w/o Multi-Expert & $\checkmark$ & $\checkmark$ & $\checkmark$ & $\times$ & $\checkmark$ & 83.8 & -6.1 & 78.7 & -8.6 \\
w/o Code Self-Correction & $\checkmark$ & $\checkmark$ & $\checkmark$ & $\checkmark$ & $\times$ & 79.8 & -10.1 & 80.0 & -7.3 \\
\bottomrule
\end{tabular}
}

\vspace{2pt}
\footnotesize
Accuracy (\%); $\Delta$: drop from Full Method; \na: not applicable.
\end{table*}

Both the three-stage structure and the stage-specific mechanisms contribute to overall effectiveness (Table~\ref{tab:ablation-stage-mechanisms}).
Removing semantic understanding or formulation causes accuracy drops of 6.1--9.1 pp on IndustryOR and 8.6--9.3 pp on NL4COP. The three-stage structure decomposes the modeling process into explicit stages for semantic understanding and mathematical construction, allowing defects in variable definitions, constraint scope, and objective structure to be identified and revised before code generation.

The stage-specific mechanisms also contribute substantially.
Removing the multi-expert mechanism costs 6.1--8.6 pp, while removing code self-correction costs 7.3--10.1 pp.
The multi-expert mechanism generates candidate formulations from diverse expert perspectives, increasing the diversity and quality of formulations that enter validation. Code self-correction iteratively refines generated code by using execution feedback to diagnose and fix errors such as syntax violations, incorrect solver API calls, and infeasible model constructions.

\subsubsection{Synergy Between Modeling Quality and Validation}

The previous analysis shows that both the modeling design and the validation framework contribute to accuracy. We now examine how they reinforce each other. Better modeling construction reduces errors at the source, gives validation formulations that are already closer to acceptance, and makes subsequent repair more focused. We use the multi-expert formulation mechanism as an example (Table~\ref{tab:multi-expert-process}).

\begin{table}[!t]
\caption{Formulation quality and validation efficiency. The multi-expert mechanism raises Pass@1 and reduces both validation iterations and execution rounds.}
\label{tab:multi-expert-process}
\centering
\footnotesize
\setlength{\tabcolsep}{2pt}
\renewcommand{\arraystretch}{1.18}
\begin{tabular*}{\columnwidth}{@{\extracolsep{\fill}}llccc@{}}
\toprule
Bench. & Variant & \makecell[c]{Pass@1\\(\%) $\uparrow$} & \makecell[c]{Valid.\\Iters $\downarrow$} & \makecell[c]{Exec.\\Rounds $\downarrow$} \\
\midrule
\multirow{2}{*}{IndustryOR}
& Full Method & \textbf{51.0} & \textbf{2.9} & \textbf{1.9} \\
& w/o Multi-Expert & 27.3 & 4.5 & 1.9 \\
\midrule
\multirow{2}{*}{NL4COP}
& Full Method & \textbf{46.0} & \textbf{3.1} & \textbf{2.9} \\
& w/o Multi-Expert & 30.7 & 4.8 & 4.1 \\
\bottomrule
\end{tabular*}
\end{table}

This interaction is reflected in how formulation quality reshapes the role of validation. Better formulation construction raises formulation Pass@1 from 27.3\% to 51.0\% on IndustryOR and from 30.7\% to 46.0\% on NL4COP. Validation then more often operates on formulations that are already close to acceptance and can focus on screening and targeted revision. It also reduces average validation iterations from 4.5 to 2.9 and from 4.8 to 3.1. On NL4COP, it further reduces code execution rounds from 4.1 to 2.9, indicating that better formulation construction reduces formulation errors that carry into code translation and execution.

\begin{figure*}[t]
\centering
\includegraphics[width=0.98\textwidth]{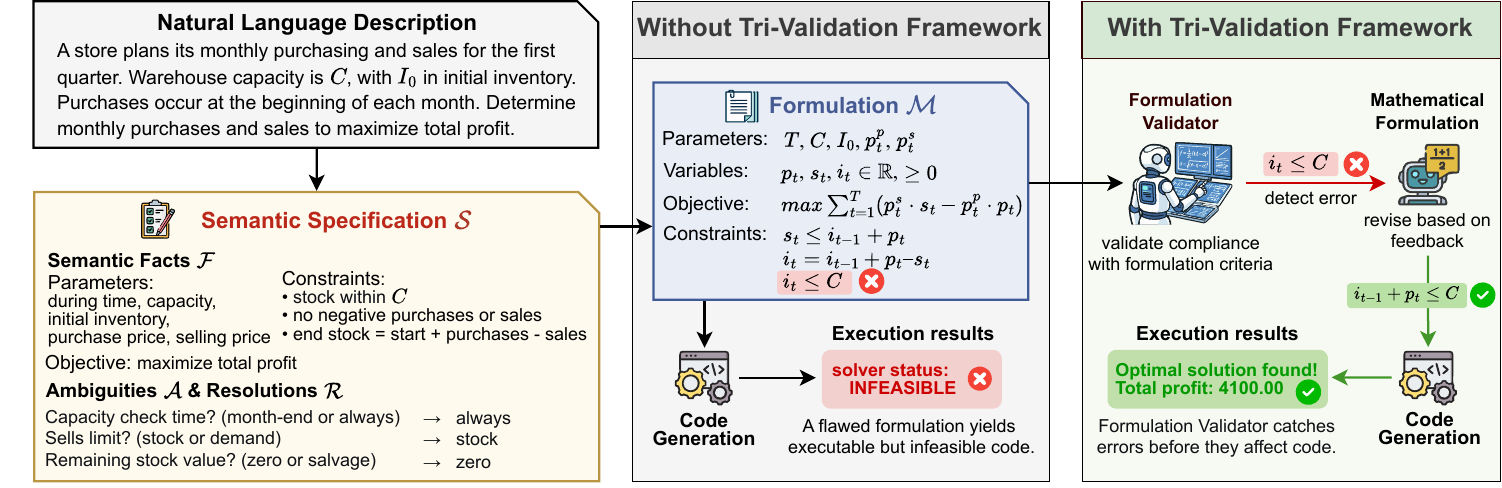}
\caption{Case study: formulation-side validation identifies an incorrect capacity constraint in $M_1$ at a stage where the defect remains interpretable, enabling a targeted revision that restores feasibility.}
\label{fig:case-study}
\end{figure*}

Fig.~\ref{fig:case-study} illustrates the same mechanism at the instance level. In the initial formulation $M_1$, the capacity constraint is applied to end-of-month inventory rather than to the inventory state after purchasing. The resulting code $C_1$ faithfully translates this incorrect formulation and therefore produces an infeasible model. In this setting, execution feedback exposes the symptom but provides limited guidance about the source of the error: infeasibility alone does not determine whether the defect comes from variable timing, constraint scope, or another formulation decision. At the formulation stage, however, the defect is still visible and amenable to targeted revision before it is translated into code and reflected in execution results.

\subsubsection{Cost of Validation}

Validation incurs additional LLM calls. Table~\ref{tab:cost-tradeoff} shows that removing all validation saves 46.8\% of tokens on IndustryOR and 33.8\% on NL4COP, but more than doubles the error rate on both benchmarks. On NL4COP, the modeling process itself is already more token-intensive because longer descriptions and greater modeling complexity demand more reasoning and more careful construction, even without validation. These additional tokens are used to evaluate intermediate modeling results and repair detected errors across the semantic, formulation, and code stages. For challenging optimization modeling tasks, reducing modeling errors and maintaining reliable, accurate modeling carry greater practical value than minimizing token usage alone.

\begin{table}[t]
\caption{Token cost and error rate with and without validation. The additional tokens spent on validation reduce the error rate by more than half on both benchmarks.}
\label{tab:cost-tradeoff}
\centering
\small
\setlength{\tabcolsep}{3pt}
\renewcommand{\arraystretch}{1.2}
\begin{tabular*}{\columnwidth}{@{\extracolsep{\fill}}lcccc@{}}
\toprule
Benchmark & \makecell[c]{Tokens\\(Full / w/o)} & \makecell[c]{Token\\Saving\\(\%)} & \makecell[c]{Error (\%)\\(Full / w/o)} & \makecell[c]{Error\\Increase\\($\times$)} \\
\midrule
IndustryOR & 54.9k / 29.2k & 46.8 & 10.1 / 21.2 & 2.1 \\
NL4COP & 62.8k / 41.6k & 33.8 & 12.7 / 30.7 & 2.4 \\
\bottomrule
\end{tabular*}

\vspace{2pt}
\footnotesize
Tokens and errors are reported as Full / w/o Validation values.
\end{table}

\section{Conclusion and Future Work}
\label{sec:conclusion}

This work presents TriVAL, a tri-validation framework for automatic optimization modeling, and NL4COP, a new benchmark with 50 problem types and 150 instances that provides a more challenging and discriminative evaluation than existing benchmarks. TriVAL introduces explicit validation at three stages of the modeling pipeline (semantic specification, mathematical formulation, and generated code), evaluating each result before errors propagate across stages. Experiments on both established benchmarks and NL4COP show that staged validation consistently improves modeling accuracy, with the largest gains on the most challenging problems. Validation of intermediate results is the key driver of these improvements, and the same validation mechanism also improves a representative existing optimization-modeling framework. These results position the tri-validation approach as an effective design principle for automatic optimization modeling and NL4COP as a challenging benchmark that clearly distinguishes the modeling capabilities of different optimization-modeling methods.

Future work will focus first on strengthening validation under more challenging problem settings. As problem complexity grows, semantic ambiguity, complex constraint interactions, and long-range dependencies make modeling errors harder to identify reliably. Since TriVAL relies on LLM-based validators, subtle constraint violations and ambiguous formulations may still challenge their assessments. A natural next step is to incorporate richer problem-aware signals, finer-grained cross-representation consistency checks, and more precise evaluation criteria to improve validation quality.

Beyond the combinatorial problems studied here, extending TriVAL to multi-objective, stochastic, robust, and dynamic optimization settings opens a broader research direction for both modeling and validation. Improving validation efficiency while maintaining strong performance is also an important practical goal. On the benchmark side, expanding NL4COP to cover these broader problem classes would provide a more comprehensive testbed for future research.


\bibliographystyle{IEEEtran}
\bibliography{references}

\end{document}